\documentclass{article}

    \usepackage[preprint]{neurips_2020}

\usepackage[utf8]{inputenc} 
\usepackage[T1]{fontenc}    
\usepackage{hyperref}       
\usepackage{url}            
\usepackage{booktabs}       
\usepackage{amsfonts}       
\usepackage{nicefrac}       
\usepackage{microtype}      

\usepackage{bbm}
\usepackage{amsmath}
\usepackage{amssymb}
\usepackage{xcolor}
\usepackage{multirow}
\usepackage{graphicx}
\usepackage{subfigure}

\usepackage{wrapfig}
\usepackage{custom_commands}
\providecommand{\pcite}[1]{{(\cite{#1})}}

\title{\methodname{}}

\author{
  Shraman Ray Chaudhuri\thanks{Work done as part of the Google AI Residency program.} \\
  Google Research \\
  Mountain View, CA \\
  \texttt{shramanrc@gmail.com} \\
  \And
  Elad Eban, Hanhan Li, Max Moroz, Yair Movshovitz-Attias \\
  Google Research \\
  Mountain View, CA \\
  \texttt{ \{elade, uniqueness, pkch, yairmov\}@google.com } 
}

\begin{document}

\maketitle

\begin{abstract}
State-of-the-art deep networks are often too large to deploy on mobile devices and embedded systems.
Mobile neural architecture search (NAS) methods automate the design of small models but state-of-the-art NAS methods are expensive to run.
Differentiable neural architecture search (DNAS) methods reduce the search cost but explore a limited subspace of candidate architectures.
In this paper, we introduce \methodname{} (\methodabbr{}), a differentiable search method that searches over a much larger set of candidate architectures.
\methodabbr{} simultaneously selects and modifies operators in the search space by applying a structured sparse regularization penalty based on the \regname{} distribution.
We show results across 3 existing search spaces, matching or outperforming the original search algorithms and producing state-of-the-art parameter-efficient models on ImageNet (e.g., 75.4\% top-1 with 2.6M params).
Using our architectures as backbones for object detection with SSDLite, we achieve significantly higher mAP on COCO (e.g., 25.8 with 3.0M params) than MobileNetV3 and MnasNet.
\end{abstract}


\section{Introduction}

Machine learning researchers have invested much effort over the last decades into feature engineering, the process of hand-crafting features for machine learning algorithms.
With the proliferation of deep learning, this process has been replaced by the manual design of larger and more complex models.

Model design requires domain expertise and many rounds of trial-and-error.
Neural architecture search (NAS) \pcite{zoph2016neural} automates this process using RL; however, searching for a new architecture can require thousands of GPU hours.
Due to the cost of prevailing NAS methods, most techniques search for an architecture over a small proxy dataset and release the discovered architecture as their contribution.
This is suboptimal---neither the proxy dataset nor the resource constraints targeted during the search could possibly address all downstream uses of this architecture.

Differentiable NAS (DNAS) \pcite{cai2018proxylessnas, liu2018darts} methods aim to alleviate this limitation by building a superset network (\textit{super-network} or \textit{search space}) and searching for useful sub-networks using gradient descent.
These super-networks are typically composed of densely connected building blocks with multiple parallel operations.
The goal of the search method is to prune connections and operations, trading representational capacity for efficiency, to fit a certain computation budget.
DNAS methods can be viewed as pruning methods with the subtle difference that they are applied on manually designed super-networks with redundant components while pruning methods \pcite{lecun1990optimal, han2015learning} are usually applied on standard models.

The canonical approach to DNAS is to select an operator from a fixed set of operators by gating their outputs and treating them as unmodifiable (black-box) units.
In this sense, DNAS has inherited some of the limitations of RL methods since they cannot dynamically change the units during optimization.
For instance, to learn the width of each layer, DNAS and RL methods typically enumerate a set of fixed-width operators, generating independent outputs for each.
This is not only computationally expensive but also a coarse way of exploring sub-networks.

We propose \methodabbr{} (\methodname{}), a search method inspired by structured pruning.
For each output feature of a layer (i.e., output channel in convolutional layer, or neuron in a dense layer), we assign a Bernoulli random variable (\textit{mask}) indicating whether that feature should be used.
We use the Logistic-Sigmoid distribution \pcite{maddison2016concrete} to relax the binary constraint and learn the masking probabilities using gradient descent, optimizing for various resource constraints such as model size or FLOPs.
We export an architecture, defined as a mapping from layers to numbers of neurons, by sampling a mask at the end of training. 

\methodabbr{} can be applied to any search space by simply inserting masks after each layer.
Our method is \textit{fine-grained} in that we search over a larger space of architectures than ordinary DNAS methods by applying masks on operator outputs as well as on intermediate layers that compose the operator.
\methodabbr{} can simultaneously select a subset of operators and modify them as well.

In some sense, DNAS has shifted the problem of architecture design to search space design.
Many DNAS works target a single metric on a single, manually-designed search space; however, each search space may come with its own merits.
This coupling between search space and algorithm makes it hard to (1) compare different search algorithms, and (2) understand the biases inherent to different search spaces \pcite{sciuto2019evaluating, radosavovic2019network}.
NAS-bench-101 \pcite{ying2019bench} addresses the former, providing a large set of architectures trained on CIFAR to evaluate RL-based NAS algorithms.
On the other hand, our method can be used to study the latter.
Since our method can easily be injected into any DNAS search space, we can characterize their bias toward certain metrics. We find that some produce models that are more Pareto-efficient for model size while others are more Pareto-efficient for FLOPs/latency.

When applied to well-known search spaces \pcite{bender2018understanding, wu2019fbnet}, \methodabbr{} matches or outperforms the original search method.
When applied on the One-Shot search space, \methodabbr{} achieve state-of-the-art small model accuracy on ImageNet (by a 2-5\% margin).
When using ImageNet-learned architectures as backbones for detection, \methodabbr{} achieves +4 mAP over mobile baselines on COCO.
When applied to commonly used ResNet models, \methodabbr{} outperforms pruning baselines.


\section{Related Work}

Neural architecture search (NAS) automates the design of neural net models with machine learning.
Early approaches \pcite{zoph2016neural, baker2016designing} train a controller to build the network with reinforcement learning (RL).
These methods require training and evaluating thousands of candidate models and are prohibitively expensive for most applications.
Weight sharing methods \pcite{brock2017smash, pham2018efficient, cai2018efficient} amortize the cost of evaluation; however, \pcite{sciuto2019evaluating} suggest that these amortized evaluations are noisy estimates of actual performance.

Of growing interest are \textit{mobile} NAS methods which produce smaller architectures that fit certain computational budgets or are optimized for certain hardware platforms.
MnasNet \pcite{tan2019mnasnet} is an RL-based method that optimizes directly for specific metrics (e.g., mobile latency) but takes several thousand GPU-hours to search.
One-shot and differentiable neural architecture search (DNAS) \pcite{bender2018understanding, liu2018darts} methods cast the problem as finding optimal subnetworks in a continuous relaxation of NAS search spaces which is optimized using gradient descent.

Our work is most closely related to probabilistic DNAS methods which learn stochastic gating variables to select operators in these search spaces.
\pcite{cai2018proxylessnas} use hard (binary) gates and a straight-through estimation of the gradient, whereas \pcite{xie2018snas, wu2019fbnet, dong2019searching} use soft (non-binary) gates sampled from the Gumbel-Softmax distribution \pcite{jang2016categorical, maddison2016concrete} to relax the discrete choice over operators.
In contrast, our method performs a fine-grained search over the set and composition of operators.
Some methods learn a single cell structure that is repeated throughout the network \pcite{dong2019searching, xie2018snas, bender2018understanding} whereas our method learns cell structures independently.

Our work draws inspiration from structured pruning methods \pcite{luo2017thinet, liu2017learning, wen2016learning}.
MorphNet \pcite{gordon2018morphnet} adjusts the number of neurons in each layer with an $\ell_1$ penalty on BatchNorm scale coefficients, treating them as gates.
\pcite{louizos2017learning} propose a method to induce exact sparsity for one-round compression.
Recent work by \pcite{mei2020atomnas} independently proposes fine-grained search with an $\ell_1$ penalty.
In contrast, we propose a stochastic method that samples sparse architectures throughout the search process.

Recent analytical works highlight the importance of search space design.
Of particular relevance is the study in \pcite{radosavovic2019network} which finds that randomly sampled architectures from certain spaces (e.g., DARTS \pcite{liu2018darts}) are superior when normalizing for certain measures of complexity.
\pcite{sciuto2019evaluating} find that randomly sampling the search space produces architectures on par with both controller-based and DNAS methods.
\pcite{xie2019exploring} suggest that the wiring of search spaces plays a critical role in the performance of sub-networks.
The success of NAS methods, therefore, can be attributed in no small part to search space design.


\section{Method}

\begin{figure*}[ht]
\vskip 0.2in
\begin{center}
\centerline{\includegraphics[width=1\columnwidth,height=0.25\columnwidth]{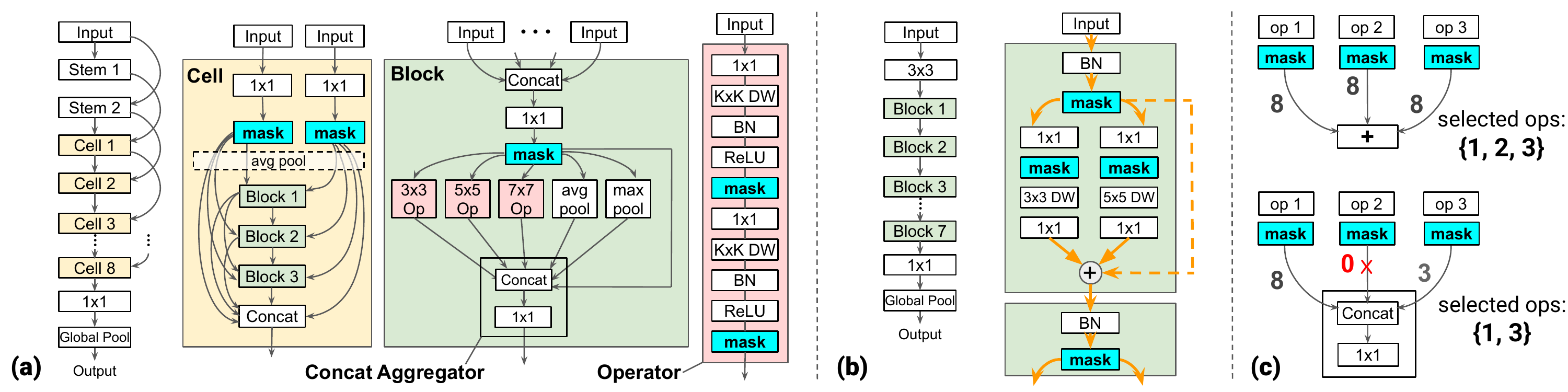}}
\caption{\small{Search Spaces and Operator Selection with \methodabbr{}. \textbf{(a)} and \textbf{(b)}: \fcell{} and \fbcell{} search spaces (resp.). ``DW" denotes depthwise convolution. Orange edges in (b) indicate tensors that must have the same shape due to the additive skip connection. \textbf{(c)}: To control the set of operators, one only needs to insert masking layers (blue) after them. Numbers next to edges indicate the number of non-zero channels in the mask. We deselect operators by sampling a zero mask. Note that the additive aggregator (top) forces the output shapes of each operator to match whereas the concat aggregator (bottom) allows arbitrary shapes and selecting a subset.}}
\label{search-space}
\end{center}
\vskip -0.2in
\end{figure*}

We search for architectures that minimize both a task loss $\Lt$ and a computational cost $\Lc$. Our approach is akin to stochastic differentiable search methods such as \pcite{xie2018snas} which formulate the architecture search problem as sampling subnetworks in a large supernetwork composed of redundant components (operators). While efficient, these methods add restrictive priors to the learned architectures: (1) the operators (e.g., a set of depthwise-separable convolutions with various widths and kernel sizes) are hand-designed and the search algorithm cannot modify them, and (2) the search algorithm is limited to selecting a \textit{single} operator for each layer in the network.

\methodabbr{} relaxes both constraints by (1) modifying operators during the search process, and (2) allowing more than one operator per layer.
To concretely illustrate the benefit of (1), we focus on the width (i.e., number of filters or neurons) of each convolution.
To modify the width during the search process, \methodabbr{} learns a sampling distribution over individual neurons in the supernetwork instead of a distribution over operators.
As a result, the operators in the learned architectures can have fine-grained, variable widths which are not limited to a pre-defined set of values.

\methodabbr{} progresses in two phases: an architecture learning phase (\SL{}) where we output an optimized architecture by minimizing both $\Lt$ and $\Lc$, followed by a retraining phase with $\Lt$ alone.
Our loss for AL is similar to sparsity-inducing regularizers \pcite{gordon2018morphnet}.
Sec. \ref{sec-lsr} describes our stochastic relaxation of $\Lc$ and sampling method, \ref{sec-sparsity} describes the masking layer and regularization penalty, and Sec. \ref{sec-s3} describes our formula for fine-grained search on existing spaces.


\subsection{Inducing Sparsity with the \regname{} Distribution} \label{sec-lsr}

Let $\bw$ be the weights of the network. We assume computational costs of the form $\Lc(\{\mathbbm{1}_{w_i \neq 0}\})$, i.e., a function of the set of indicators for whether each weight is nonzero.
FLOPs, size, and latency can be expressed exactly or well-approximated in this form.
The \SL{} objective is then:

\begin{equation}
\min_{\bw}\{\Lt(\bw) + \lambda \Lc(\{\mathbbm{1}_{w_i \neq 0}\})\} \label{eq-original}
\end{equation}

We refer to $\lambda$ as the \textit{regularization strength}.
Since $\mathbbm{1}_{w_i \neq 0}$ has zero gradient when $w_i \neq 0$, we cannot minimize $\Lc$ with gradient descent.
Instead, we formulate the problem as a stochastic estimation task. Let $\bm$ denote a binary mask to be applied on $\bw$, where $m_i \sim Bern(\pi_i)$ are independent Bernoulli variables.
We minimize the usage of $w_i$ by minimizing the probability $\pi_i$ that the mask is 1 so we can safely prune $w_i$.
Our sampled architectures are defined by $\bm$.
By substituting $\{\mathbbm{1}_{w_i \neq 0}\}$ with $\bm$, our objective becomes:
\begin{equation}
    \min_{\boldsymbol{\pi}, \bw}\{\mathbbm{E}_{\bm\sim Bern(\boldsymbol{\pi})} [\Lt(\bw \odot \bm) + \lambda \Lc( \bm ) ]\}
    \label{eq-stochastic}
\end{equation}
where $\odot$ denotes element-wise product.
Unless otherwise specified, all expectations are taken w.r.t. $\boldsymbol{\pi}$ and we drop the subscript on $\mathbbm{E}$ for brevity.
We can estimate the gradient w.r.t. $\boldsymbol{\pi}$ with black-box methods, e.g., perturbation methods \pcite{spall1992multivariate} or the log-derivative trick \pcite{williams1992simple}; however, these estimators generally suffer from high variance.
Instead, we relax $m_i$ with a continuous sample from the \regname{} distribution:
\[
\hemm_i = \text{Sigmoid}((\log(\frac{\hpi_i}{1-\hpi_i}) + \ell)/\tau)
\]
where  $\; \ell \sim \text{Logistic}(0, 1)$.
The \regname{} distribution is the binary case of the Gumbel-Softmax (a.k.a. Concrete) distribution \pcite{jang2016categorical, maddison2016concrete}.
As $\tau \rightarrow 0$, $\hemm_i$ approaches $\{0, 1\}$ with probability $\{1-\hpi_i, \hpi_i\}$ respectively.
Factoring out logistic noise as a parameter-free component allows us to back-propagate through the mask and learn $\hpi_i$ with gradient descent.
The resulting gradient estimator has lower variance than black-box methods \pcite{maddison2016concrete}.
We optimize $\nu = \log(\frac{\hpi_i}{1-\hpi_i})$ in practice for numerical stability.

Our stochastic relaxation allows us to better model the sparsity of the learned architectures during the search phase than deterministic relaxations.
To illustrate, consider the common deterministic approach to relax $\Lc(\{\mathbbm{1}_{w_i \neq 0}\})$ with an $\ell_p$ norm where $p > 0$ \pcite{wen2016learning, gordon2018morphnet, mei2020atomnas}.
In this case, the weights can be far from $\{0, 1\}$ during training, which can be problematic if the network relies on the information encoded in these pseudo-sparse weights to make accurate predictions.
Instead, we want to simulate real sparsity during AL.
Other deterministic methods apply a saturating nonlinearity (e.g., sigmoid or softmax) to force values close to $\{0, 1\}$ \pcite{liu2018darts}.
However, these functions suffer from vanishing gradients at extrema: once a weight is close to zero, it remains close to zero.
This limits the number of sparse networks explored during AL.
In contrast, our sampled mask is close to $\{0, 1\}$ at all times at low $\tau$, which forces the network to adapt to sparse activations, and the mask can be non-zero even as $\hpi$ approaches 0, which allows the network to visit diverse sparse states during AL.


\subsection{Group Masking and Regularization} \label{sec-sparsity}

%
As neurons are pruned during the search process, we can prune downstream dependencies as well. We group each neuron and its downstream dependencies by sharing a single mask across all their elements. To illustrate, consider the weight matrices of two 1x1 convolutions $\bf A$ and $\bf B$ below, where the output of $\bf{A}$ is fed into $\bf{B}$. If neuron {\color{orange} $\mathbf{a_{2, \bullet}}$} $\approx \mathbf{0}$, then {\color{blue} $\mathbf{b_{\bullet, 2}}$} can be pruned and vice versa. Therefore, all elements in $\mba$ and $\mbb$ share a scalar mask $m_i$.
\begin{equation*}
\bf{A} =
\begin{bmatrix}
 a_{1,1} &  a_{1,2} & \cdots \\
\bf \color{orange} a_{2,1} & \bf \color{orange} a_{2,2} & \bf \bf \color{orange} \cdots \\
\vdots  & \vdots  & \ddots
\end{bmatrix}
\quad
\bf{B} =
\begin{bmatrix}
 b_{1,1} & \bf \color{blue} b_{1,2} & \cdots \\
 b_{2,1} & \bf \color{blue} b_{2,2} & \cdots \\
\vdots  & \bf \color{blue} \vdots  & \ddots
\end{bmatrix}
\end{equation*}

This row-column grouping can be implemented conveniently by applying a separate mask on each channel of the activations produced by each convolution and fully-connected layer.
This allows us to encapsulate all architecture learning meta-variables ($\boldsymbol{\hpi}$) in a drop-in layer which can easily be inserted in the search space.

Let $\LL_i^c$ be the contribution of $\mba$ and $\mbb$ to the total cost $\Lc$.
To encourage sparsity, we can either regularize the mask ($m_i$) or the distribution parameters ($\pi_i$).
As $\tau \rightarrow 0$, the former penalizes the cost of sampled architectures while the latter penalizes the expected cost.
In our example above, the sampled and expected costs (in number of parameters) are:
\begin{align}
    \LL_i^c &\approx m_i \cdot (||\mba||_0 + ||\mbb||_0) \\
    \mathbbm{E}[\LL_i^c] &\approx \pi_i \cdot (||\mba||_0 + ||\mbb||_0)
\end{align}

Note however that $||\mba||_0$ and $||\mbb||_0$ are dynamic quantities: as inputs to $\bf A$ and outputs of $\bf B$ are masked out by adjacent masking layers, $||\mba||_0$ and $||\mbb||_0$ decrease as well.
To capture this dynamic behavior, we apply our differentiable relaxation from Sec. \ref{sec-lsr} again to approximate $||\mba||_0$ and $||\mbb||_0$.
Let $\hmain_j$ and $\hmbout_k$ be per-channel masks on inputs to $\bf A$ and outputs of $\bf B$.
The sampled and expected costs are then:
\begin{align}
    \LL_i^c &\approx \hemm_i \cdot (\sum_j \hmain_j + \sum_k \hmbout_k)\label{objective-sampled} \\
    \mathbbm{E}[\LL_i^c] &\approx \hpi_i \cdot (\sum_j \hpain_j + \sum_k \hpbout_k)\label{objective-expected}
\end{align}
We observe that minimizing (\ref{objective-expected}) is more stable than minimizing (\ref{objective-sampled}).
We use (6) for our results in Sec.~4, and scale $\LL_i^c$ appropriately for different costs such as FLOPs. 

After \SL{}, we export a single architecture, defined as a mapping from each convolution layer to its expected number of neurons under the learned distribution parameters $\boldsymbol{\pi}$.
In our example above, convolution $\bf A$ would have $\floor{\sum_i \pi_i}$ neurons in the exported architecture.


\subsection{Fine-Grained Search} \label{sec-s3}

To apply our method to DNAS search spaces, we simply insert masking layers after convolution layers as illustrated in Fig. \ref{search-space}.
We run our search algorithm on the One-Shot and FBNet search spaces.
The One-Shot search space is composed of a series of \textit{cells} which are in turn composed of densely connected \textit{blocks}.
Each block consists of several operators, each of which applies a series of convolutions on the blocks' inputs.
Similarly, the FBNet search space is composed of \textit{stages} which are in turn composed of blocks.
The outputs of the operators are added together.
We refer the reader to \pcite{bender2018understanding, wu2019fbnet} for more details.

DNAS methods generally gate the operator outputs directly to select a single operator with, e.g., a softmax a layer.
In contrast, our architectures can have more than one operator per layer.
Operators are removed from the network by learning to sample all-zero masks on the operator's output or the output of any of its intermediate activations.
This process is illustrated in Fig. \ref{search-space}(c) -- note that \methodabbr{} can select between 0 and all operators in each block.
Since our method simultaneously optimizes for the set of operators and their widths, the space of possible architectures which we search is an exponentially larger superset of the original search space.

\methodabbr{} matches the performance of the original search algorithms when applied to the original One-Shot and FBNet search spaces \textbf{with no modifications}.
However, these search spaces are designed for coarse-grained search (operator-level sampling).
We propose two minor modifications to the search space to take full advantage of \textit{fine-}grained search.
Importantly, these modifications \textbf{do not} improve the accuracy of the architectures in and of themselves; they only give more flexibility for fine-grained search and reduce the runtime of the search phase.

\textbf{Concat Aggregator.}
By adding operator outputs, we enforce all output dimensions to match during \SL{}.
This restricts fine-grained search in that each operator must have the same output shape.
Instead, we can concatenate them and pass them through a 1x1 convolution (\textit{concat aggregator}), which is a generalization of the additive aggregator.
The benefits are two-fold: (1) operator outputs can have variable sizes, and (2) \methodabbr{} can learn a better mixing formula for operator outputs.
In practice, we observe that the concat aggregator works better on the One-Shot search space when targeting model size and the additive aggregator works better on FBNet when targeting FLOPs.

\textbf{Removing Redundant Operators.}
To explore various architectural hyperparameter choices, coarse-grained NAS methods enumerate a discrete set of options.
For instance, to learn whether a convolution in a given block should have 16 or 32 filters would require including two separate weight tensors in the set of options.
Not only is this computationally inefficient -- scaling both latency and memory with each additional operator -- but the enumeration may not be granular enough to contain the optimal size. 
In contrast, fine-grained search can shrink the 32-filter convolution to be functionally equivalent to the 16-filter convolution; therefore, we only need to include the former.
In practice, this results in a 3x reduction in the number of operators in the FBNet search space and a 2.5x reduction in search runtime, with no loss of quality in the learned architectures.


\section{Results}

We use TensorFlow \pcite{abadi2016tensorflow} for all our experiments.
Our algorithm takes $8$ hours to search and $36$ hours to retrain on ImageNet using a single 4x4 (32-core) Cloud TPU.


\subsection{\methodabbr{} on One-Shot Search Space}\label{sec-oneshot}

\begin{figure}[ht!]
\begin{center}
    \centering
    \begin{minipage}{0.45\textwidth}
        \centering
        \includegraphics[width=0.9\textwidth]{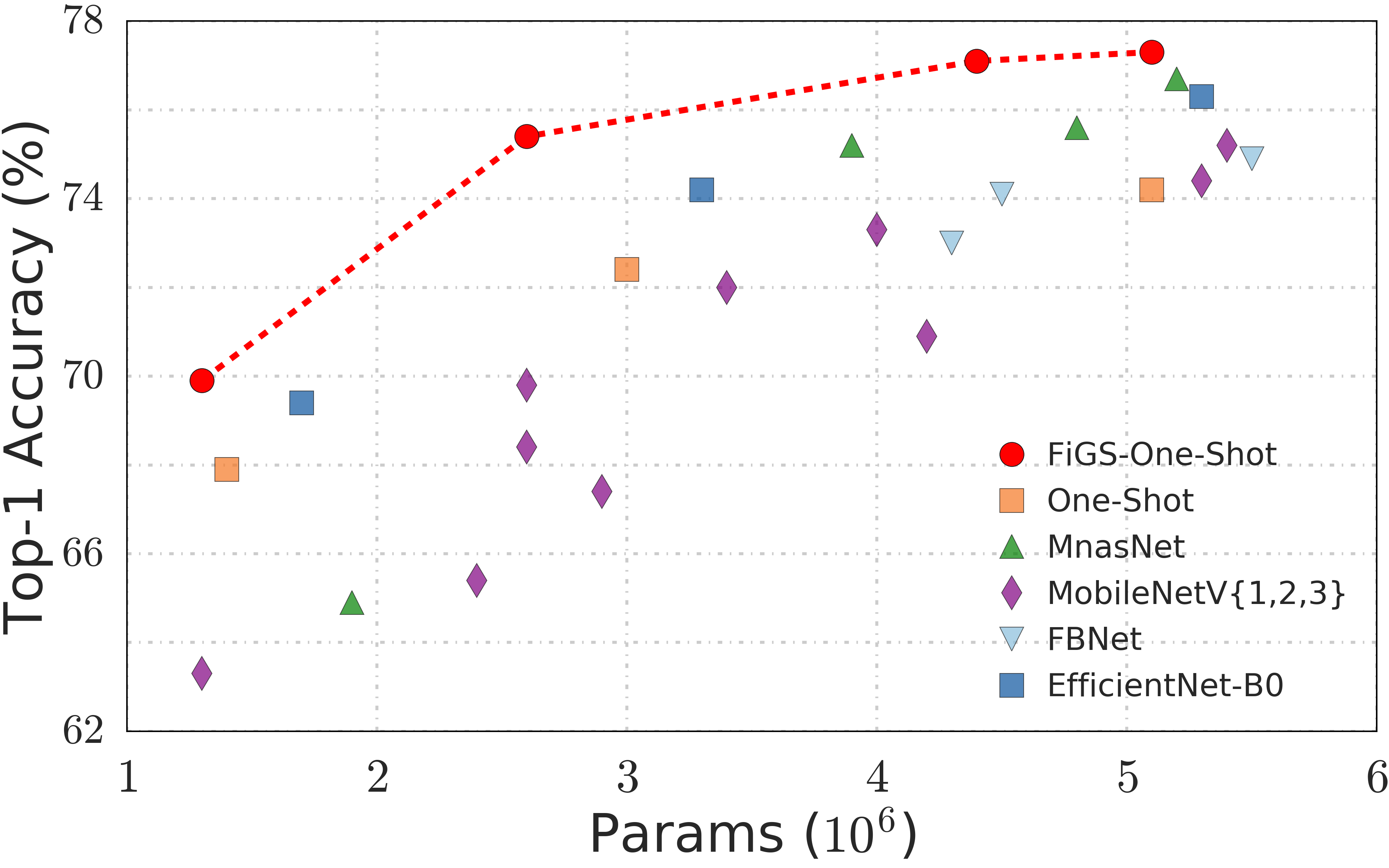}
    \end{minipage}\hfill
    \begin{minipage}{0.45\textwidth}
        \centering
        \includegraphics[width=0.9\textwidth]{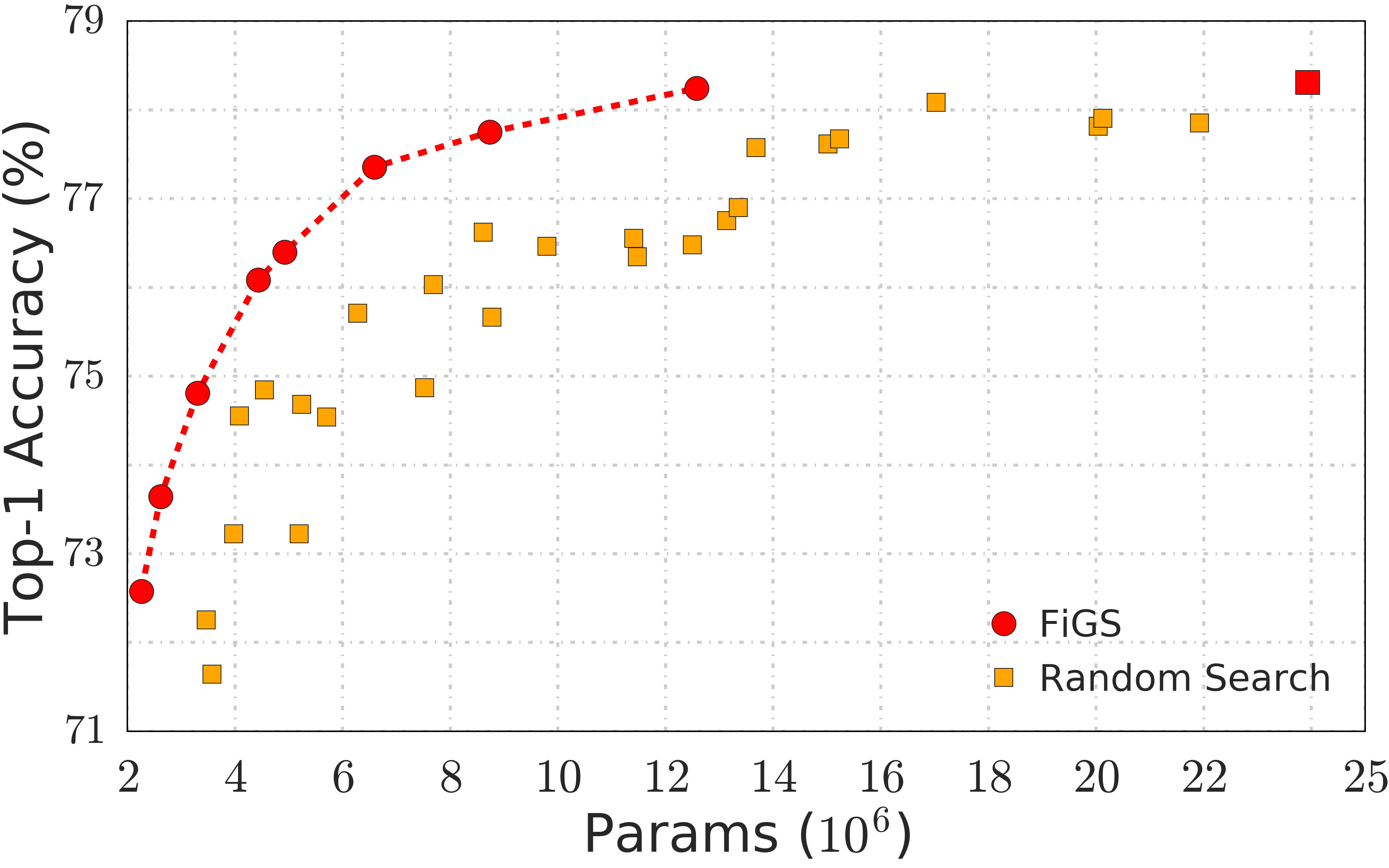}
    \end{minipage}

\caption{\small{\textbf{Left:} Model size vs Accuracy for state-of-the-art mobile architectures on ImageNet. The architectures learned by \fcell{} produce SOTA results. \textbf{Right:} \methodabbr{} vs. Random Search. We sample 30 architectures (yellow) from the One-Shot space with random subsets of operators and width multiplier $\in [0.25\times, 1.0\times]$. The right-most red point is the supernetwork. \methodabbr{} outperforms random search.}}
\label{imagenet-fig}
\end{center}
\vskip -0.1in
\end{figure}

In this section, we evaluate our search algorithm on the One-Shot search space \pcite{bender2018understanding} to find efficient architectures for ImageNet classification \pcite{russakovsky2015imagenet}.
To compare against their results, we target model size.
We use the same search space instantiation --- 8 cells, 4 blocks per cell, separable convolutions, and downsampling every 2 cells.
We merge the outputs of each path with a concat aggregator.
Despite increasing the number of parameters, the concat aggregator does not increase the base accuracy of the supernetwork.
The search space is illustrated in Fig. \ref{search-space}(a) --- note that we apply masks on operator outputs as well as on individual convolutions that compose the operator.
Our reproduction of their supernetwork matches their published results.

The mask-logits variables $\nu$ are initialized to $2.5$ ($\hpi \approx 0.92$).
We set $\tau = 0.001$ without annealing and use our relaxation of $\hemm$ for both forward and backward passes.
To learn architectures of different sizes, we vary the regularization coefficient $\lambda$.
For \SL{}, we train for 100 epochs using ADAM with batch size 512 and learning rate 1.6 decayed by 0.5 every 35 epochs.
For retraining, we use the same settings, except we train until convergence (400 epochs) and double the batch size and learning rate (1024 and 3.2, resp.) for faster convergence \pcite{smith2017don}.

\begin{table*}
    \centering
    \small
    \resizebox{0.70\textwidth}{!}{
        \begin{tabular}{l|c|cc}
        \toprule [0.15em]
        Model &  Top-1 Acc. & \#Params \\
        \midrule [0.1em]
        \textbf{\fcell{}-A} \; ($\lambda=1.2 \times 10^{-6}$)  & \textbf{69.9$\pm$0.1\%} & \textbf{1.3$\pm$0.02M} \\
        One-Shot-Small \pcite{bender2018understanding} & 67.9\% & 1.4M \\
        MnasNet-Small \pcite{tan2019mnasnet} & 64.9\% & 1.9M \\
        MobileNetV3-Small-0.75 \pcite{howard2019searching} & 65.4\% & 2.4M \\

        \midrule
        \textbf{\fcell{}-B} \; ($\lambda=5 \times 10^{-7}$)  & \textbf{75.0$\pm$0.5\%} & \textbf{2.7$\pm$0.06M} \\
        MobileNetV3-Small-1.0 \pcite{howard2019searching} & 67.4\% & 2.9M \\
        One-Shot-Small \pcite{bender2018understanding} & 72.4\% & 3.0M \\
        MnasNet-65 \pcite{tan2019mnasnet} & 73.0\% & 3.6M \\
        FBNet-A \pcite{wu2019fbnet} & 73.0\% & 4.3M \\

        \midrule
        \textbf{\fcell{}-C} \; ($\lambda=3 \times 10^{-7}$)  & \textbf{77.1$\pm$0.03\%} & \textbf{4.4$\pm$0.04M} \\
        AtomNAS-B \pcite{mei2020atomnas} & 75.5\% & 4.4M \\
        One-Shot-Small \pcite{bender2018understanding} & 74.2\% & 5.1M \\
        EfficientNet-B0 \pcite{tan2019efficientnet} & 76.3\% & 5.3M \\
        MobileNetV3-Large-1.0 \pcite{howard2019searching} & 75.2\% & 5.4M \\

        \bottomrule[0.15em]
        \end{tabular}
    }
    \caption{\small{Comparison with modern mobile classification architectures and DNAS methods on ImageNet. \methodabbr{} produces the smallest and most accurate models in each category, and significantly outperforms the One-Shot baseline. Error bars were computed by running \SL{} 6 times with the same $\lambda$, exporting a single architecture after each run as described in Sec. \ref{sec-sparsity}, and retraining from scratch.}}
    \vspace{-0pt}
    \label{imagenet-table}
\end{table*}

Fig. \ref{imagenet-fig} shows the performance of \methodabbr{} against the One-Shot algorithm and random search.
Table \ref{imagenet-table} shows our results compared with other mobile classification models.
Our search algorithm outperforms both random search and the One-Shot search algorithm, and achieves state-of-the-art top-1 accuracy in the mobile regime across several mobile NAS baselines, outperforming EfficientNet-B0 and MobileNetV3.
Our search time is comparable with other DNAS methods and significantly faster than MnasNet, which supplies the base network for MobileNetV3 and EfficientNet.
Note that our full search space has 78.5\% top-1 accuracy on ImageNet which is an upper bound on the performance of our sub-networks.
Although this upper bound is well below state-of-the-art ImageNet accuracy, we are still able to produce state-of-the-art small-models.

\begin{table}
\begin{center}
\small{

\begin{tabular}{cc}

\begin{tabular}{|l|c|c|}
\hline
Model & Acc. & MAdds \\
\hline\hline
{\fbcell{}-B} & 72.2 & 295M \\
FBNet-B & 72.3 & 295M \\
\hline
\fbcell{}-C & 73.5 & 385M  \\
FBNet-C & 73.3 & 385M \\
\hline
\end{tabular} &

\begin{tabular}{|l|c|c|c|c|}
\hline
Target & Search Space  & Acc. & Params & MAdds \\
\hline\hline
\multirow{2}{*}{Params} & \fcell{} & \textbf{77.1} & \textbf{4.4M} & -- \\
& \fbcell{}  & 74.7 & 5.2M & -- \\
\hline
\multirow{2}{*}{FLOPs} & \fcell{} & 68.0 & -- & 400M \\
& \fbcell{} & \textbf{74.0} & -- & \textbf{400M} \\
\hline
\end{tabular}

\end{tabular}
}
\end{center}
\caption{\small{\textbf{Left}: \methodabbr{} on the FBNet search space. Using our drop-in sampling layer we are able to effectively search the FBNet space, and match the performance of models found by \pcite{wu2019fbnet}. \textbf{Right}: Search spaces may have intrinsic biases from their manual construction. We find that models produced from the One-Shot space are parameter efficient while those from the FBNet space are FLOPs efficient.}}
\vspace{-20pt}
\label{fbnet-imagenet}
\end{table}


\subsection{Comparing Search Spaces}\label{sec-comparing}

We investigate whether certain search spaces are suited for particular computational costs and provide evidence in favor.
A rigorous study would require enumerating and evaluating all searchable subnetworks on each space, which is infeasible.
Instead, \pcite{sciuto2019evaluating, radosavovic2019network} study the efficiency of search spaces by randomly sampling architectures.
This analysis is useful in determining the inherent advantages of each search space independently of the search algorithm being used. However, search algorithms may be biased toward particular sub-spaces of architectures based on the specific cost targeted during search \pcite{gordon2018morphnet} and uniform sampling may not capture this bias.
Therefore, in addition to random sampling, it may be useful to compare search spaces via the performance of a search algorithm under different cost objectives.

We investigate with FBNet \pcite{wu2019fbnet} since its construction significantly differs from the One-Shot search space and similar constructions are used in other works \pcite{howard2019searching, mei2020atomnas}.
We use FLOPs as a second metric of interest.
To make a meaningful comparison between search spaces, we first verify that \methodabbr{} matches the performance of FBNet search as shown in Table~\ref{fbnet-imagenet} (left).\footnote{\label{footnote-fbnet} FBNet-\{B, C\} and \fbcell{}-\{B, C\} were evaluated using our re-implementation of their training code.
}
We then run \methodabbr{} with both FLOPs and size costs on One-Shot and FBNet search spaces as shown in Table~\ref{fbnet-imagenet} (right). \methodabbr{} finds more parameter-efficient networks in the One-Shot search space and FLOPs-efficient networks in the FBNet search space by significant margins.


\begin{wrapfigure}{r}{0.5\textwidth}
  \vspace{-40pt}
  \begin{center}
    \includegraphics[width=0.4\textwidth]{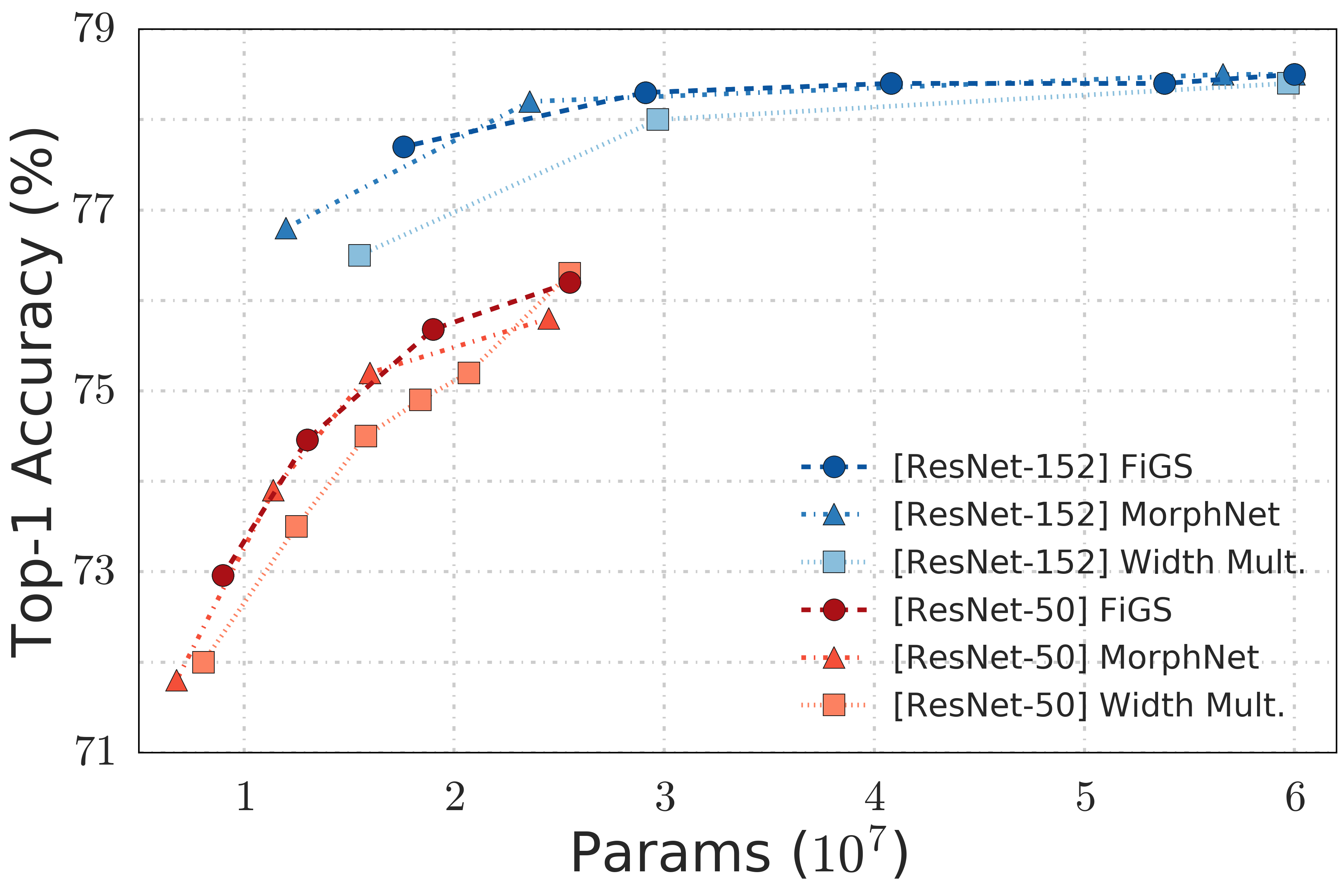}
  \end{center}
   \vspace{-10pt}
  \caption{\small{FiGS vs. architecture compression baselines on ResNet-\{50, 152\}. FiGS outperforms width multiplier and performs on par with MorphNet.}}
  \label{resnet}
   \vspace{-12pt}
\end{wrapfigure}


\subsection{\methodabbr{} on ResNet Search Space}\label{sec-resnet}

We compare the performance of \methodabbr{} with (1) width multiplier, a commonly used compression heuristic that uniformly scales down the number of filters in each layer \pcite{howard2017mobilenets} and (2) MorphNet, a deterministic model compression technique which uses $\ell_1$ regularization to induce sparsity \pcite{gordon2018morphnet}.
We use MorphNet as a baseline since it can target various computational costs and \pcite{mei2020atomnas} use a similar $\ell_1$ technique.

Fig. \ref{resnet} shows our results on ResNet-50 and ResNet-152 on ImageNet.
On both networks, \methodabbr{} outperforms width multiplier and performs on par with MorphNet.


\subsection{Mobile Object Detection}\label{sec-detection}

In this section, we demonstrate the performance of our ImageNet-learned \fcell{} architectures as backbones for mobile object detection, using the SSDLite meta-architecture \pcite{sandler2018mobilenetv2} designed for small models.
We connect the output of cell 5 (stride 16) to the first layer of the feature extractor and output of the final 1x1 before global pool (stride 32) to the second layer.
We compare against MobileNets and MnasNet which both use SSDLite.

\begin{table}
\begin{center}
 \resizebox{0.65\columnwidth}{!}{

\begin{tabular}{|l|c|c|c|c|c|}
\hline
Backbone & Params & mAP & mAP$_s$ & mAP$_m$ & mAP$_l$ \\
\hline\hline
\fcell{}-Small & 1.91M & 19.1 & 2.6 & 15.4 & 37.5 \\
MobileNetV3-Small & 1.77M & 14.9 & 0.7 & 5.6 & 28.0 \\
\hline
\fcell{}-Large & 3.02M & 25.8 & 4.4 & 24.1 & 47.8 \\
MobileNetV3-Large & 3.22M & 21.8 & 1.9 & 12.7 & 40.7 \\
MnasNet-A1 & 4.90M & 23.0 & 3.8 & 21.7 & 42.0 \\
\hline
\end{tabular}
}
\label{detection}
\end{center}
\caption{\small Mobile object detection on COCO 2017 test-dev with SSDLite meta-architecture. \fcell{}-Large outperforms both MobileNetV3-Large and MnasNet-A1 with fewer params.}
\vspace{-10pt}
\end{table}

Our results are shown in Table \ref{detection}.
We achieve a +4 mAP margin over MobileNetV3.
Note that instead of transferring ImageNet-learned architectures, we could also apply our search method to learn the backbone directly on the detection dataset, using differentiable relaxations of search spaces designed for detection such as \pcite{chen2019detnas}.
This would likely produce more efficient architectures and is left as future work.


\subsection{On Convergence and Reducing Runtime}\label{sec-convergence}

\begin{wraptable}{r}{0.5\textwidth}
\vspace{-20pt}
\begin{center}

\small{
\begin{tabular}{|l|c|c|c|}
\hline
Target Size & $\lambda$ & Epochs (\SL{}) & Acc. \\
\hline\hline
\multirow{3}{*}{2M params} & $7\times10^{-7}$ & 100 & 71.8\% \\
& $12\times10^{-7}$ & 40 & 71.5\% \\
& $20\times10^{-7}$ & 20 & 70.4\% \\
\hline
\multirow{3}{*}{5M params} & $3\times10^{-7}$ & 100 & 76.4\% \\
& $5\times10^{-7}$ & 40 & 76.2\% \\
& $9\times10^{-7}$ & 20 & 75.8\% \\
\hline
\end{tabular}
}

\caption{\small{Effects of regularization strength and search budget on final model accuracy.
Early stopping allows 2.5x-5x saving in \SL{} time with minimal accuracy drop.
}}\vspace{-10pt}

\label{convergence}
\end{center}
\end{wraptable}

We explore the limits of reducing the sample-complexity of the architecture learning phase.
Given a target model size, we explore the trade-off between running for longer with a weak $\lambda$ and converging faster with strong $\lambda$.
We demonstrate with two different target sizes (2M and 5M params).
The results are shown in Fig. \ref{convergence}.
In both cases, we can truncate \SL{} to 40 epochs with negligible drop in accuracy, reducing the runtime of our search by 2.5x.
Searching for only 20 epochs reduces model quality by 0.5-1\% but results in a 2x speedup over the One-Shot method while still producing better models.


\section{Conclusion}

We present a fine-grained differentiable architecture search method which stochastically samples sub-networks and discovers well-performing models that minimize resource constraints such as memory or FLOPs.
While most DNAS methods select from a fixed set of operations, our method modifies operators during optimization, thereby searching a much larger set of architectures.
We demonstrate the effectiveness of our approach on two contemporary DNAS search spaces \pcite{wu2019fbnet, bender2018understanding} and produce SOTA small models on ImageNet.
While most NAS works focus on FLOPs or latency, there is significant practical benefit for low-memory models in both server-side and on-device applications.

\methodabbr{} can be applied to any model or search space by inserting a mask-sampling layer after every convolution.
Due to its small search cost, our method can learn efficient architectures for any task or dataset on-the-fly.

\clearpage

\section*{Broader Impact}
Deep models have been doubling in size every few months since 2012, and have a large carbon footprint, see~\pcite{Strubell_2019, schwartz2019green}.
Moreover state-of-the-art models are often too large to deploy on low-resource devices limiting their uses to flagship mobile devices that are too expensive for most consumers.
By automating the design of models that are lightweight and consume little energy, and doing so with an algorithm that is also lightweight and adaptive to different constraints, our community can make sure that the fruits of ML/A.I. are shared more broadly with society, are not limited to the most affluent, and do not become a major contributor to climate change.

\section*{Contributions}
Shraman led the research, ran most of the experiments, and wrote most of the paper.
Yair helped with writing and provided valuable feedback through code review.

Elad and Yair jointly proposed the idea to apply structured pruning for DNAS.
Hanhan, Shraman, and Yair jointly proposed the idea to apply Gumbel-Softmax for fine-grained search.
Yair developed the Logistic-Sigmoid regularizer, and the method matured through continuous discussion between Elad, Shraman, and Yair.

Max ran experiments on object detection and provided critical engineering help.
Elad and Hanhan ran experiments on ResNet.


\clearpage

\section*{Appendix}

\subsection*{Hyperparameters and Training Details}

We use the following training setup for One-Shot, FiGS-One-Shot, FBNet, and FiGS-FBNet:

\begin{itemize}
    \item Batch size 512 and smooth exponential learning rate decay initialized to 1.6 and decayed by 0.5 every 35 epochs.
    \item Moving average decay rate of 0.9997 for BatchNorm eval statistics and eval weights.
    \item ADAM optimizer with default hyperparameters: $\beta_1 = 0.9, \beta_2 = 0.999, \epsilon = 1.0$.
    \item Weight decay with coefficient $1.7 \times 10^{-5}$.
    \item Standard ResNet data augmentation \pcite{he2016deep}: random crop, flip, color adjustment.
\end{itemize}

We use the same setup for our ResNet results in Sec. 4.3, except we set the LR schedule to be closer to \pcite{he2016deep}: initializing to 0.64 and smoothly decaying by 0.2 every 30 epochs.

We use the above training setup for both \SL{} and retraining, with the exception that we retrain until convergence. To accelerate retraining, we double the batch size and learning rate (1024 and 3.2, respectively) as per \pcite{smith2017don}. This does not improve the accuracy of our models. We do not tune hyperparameters of our learned architectures.

We provide regularization strengths ($\lambda$) for \methodabbr{}-One-Shot in Table 1.
Regularization strengths for \methodabbr{}-FBNet-(B,C) are ($2\times10^{-9}, 1.3\times10^{-9}$) respectively.

To find an appropriate order-of-magnitude for $\tau$, we log-scale searched (once) for $\tau \in \{1.0, 0.1, 0.01, 0.001, 0.0001, 10^{-5}\}$ on \methodabbr{}-One-Shot.
We found that setting $0.01 \geq \tau \geq 0.0001$ produced indistinguishable results, and fixed $\tau = 0.001$ for all experiments.

Recent works \pcite{lindauer2019best, mei2020atomnas} mention the use of special techniques in NAS works. To be explicit, we do \textbf{not} use these special techniques in training our models:
\begin{itemize}
    \item Squeeze-Excite layers.
    \item Swish activation.
    \item CutOut, MixUp, AutoAugment, or any other augmentation not explicitly listed in our training setup.
    \item Dropout, DropBlock, ScheduledDropPath, Shake-Shake or any other regularization not explicitly listed in our training setup above.
\end{itemize}

Without these techniques, we are able to outperform state-of-the-art architectures like EfficientNet-B0 which use some of these techniques.
Given the results of \pcite{mei2020atomnas}, we are optimistic that applying techniques like Squeeze-Excite, Swish, and AutoAugment can further increase the Pareto-efficiency of our networks, but that is outside the scope of this work.

All experiments (including One-Shot, FBNet, MorphNet baselines) were run on the same hardware (32-core Cloud TPU) using TensorFlow.

\subsection*{Search Space Details}

\begin{figure}[ht]
\vskip -0.1in
\begin{center}
\centerline{\includegraphics[width=0.4\columnwidth]{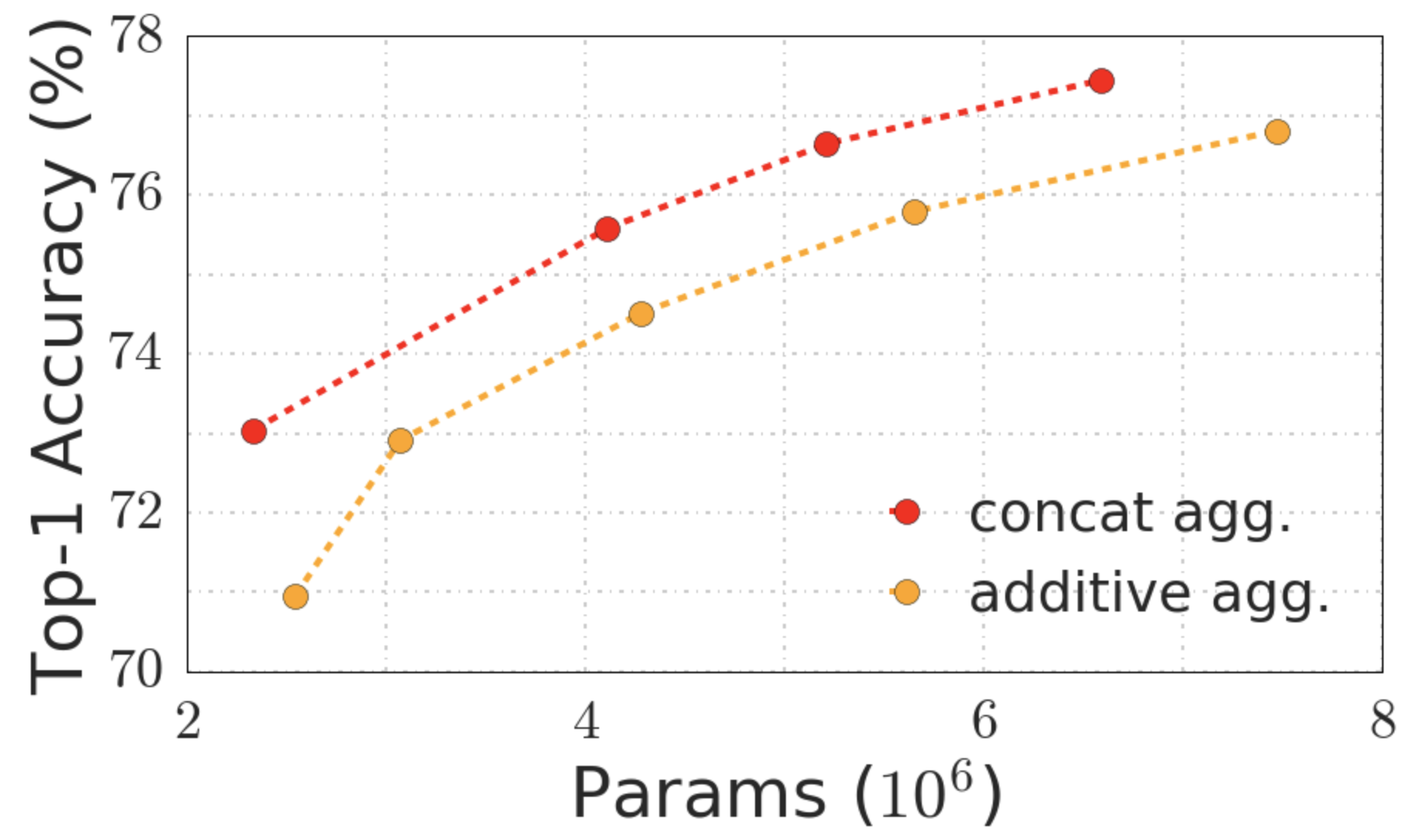}}
\caption{\small{Effect of additive vs. concat aggregator on fine-grained search on the One-Shot search space. The degree of freedom in setting the channel count of operator outputs allows \methodabbr{} to learn better architectures.}}
\label{ablation}
\end{center}
\vskip -0.2in
\end{figure}

For FiGS-One-Shot, we use the same search space instantiation presented in \cite{bender2018understanding} (sec 3.4) for ImageNet --- 8 cells, 4 blocks per cell, separable convolutions, and downsampling with stride=2 average pooling every 2 cells.
We use a base width ($F$) of 64 filters.
We verify our search space implementation by reproducing their ``All On" results in Table 1.
To assist with \textit{fine-grained} search, we make one modification, as mentioned in Sec. 3.3: we combine operator outputs by concatenating them and passing through a 1x1 convolution (instead of adding) to decouple their output dimensions.
The extra 1x1 convolution does not increase the accuracy of the supernetwork or learned architectures in and of itself.
As shown in Fig.~\ref{ablation}, the concat aggregator helps FiGS produce better architectures.

For FiGS-FBNet, we do not include group convolutions in our set of operators so we only compare against FBNet-\{B,C\} which also do not include group convolutions.




\subsection*{Miscellany}

The multiple points for EfficientNet-B0 in Fig \ref{imagenet-fig} were generated by applying a uniform width multiplier $\in \{0.5, 0.75, 1.0\}$.


\begin{table*}
    \vskip 0.1in
    \centering
    \resizebox{0.9\textwidth}{!}{                                                  
        \begin{tabular}{l|c|cc}
        \toprule [0.15em]
        Model &  Top-1 Acc. & \#Params &  Ratio-to-Ours \\
        \midrule [0.1em]                                           
\textbf{\fcell{}-A} \; ($\lambda=1.2 \times 10^{-6}$)  & \textbf{69.9$\pm$0.1\%} & \textbf{1.3$\pm$0.02M} & 1.0x \\
        One-Shot-Small \pcite{bender2018understanding} & 67.9\% & 1.4M & 1.1x \\
        MnasNet-Small \pcite{tan2019mnasnet} & 64.9\% & 1.9M & 1.5x \\
        MobileNetV3-Small-0.75 \pcite{howard2019searching} & 65.4\% & 2.4M & 1.8x \\
        
        \midrule
        \textbf{\fcell{}-B} \; ($\lambda=5 \times 10^{-7}$)  & \textbf{75.0$\pm$0.5\%} & \textbf{2.7$\pm$0.06M} & 1.0x \\
        MobileNetV2-0.75x \pcite{sandler2018mobilenetv2} & 69.8\% & 2.6M & 1.0x \\
        MobileNetV3-Small-1.0 \pcite{howard2019searching} & 67.4\% & 2.9M & 1.1x \\
        One-Shot-Small \pcite{bender2018understanding} & 72.4\% & 3.0M & 1.2x \\
        MobileNetV2-1.0x \pcite{sandler2018mobilenetv2} & 72.0\% & 3.4M & 1.3x \\
        MnasNet-65 \pcite{tan2019mnasnet} & 73.0\% & 3.6M & 1.4x \\
        AtomNAS-A \pcite{mei2020atomnas} & 74.6\% & 3.9M & 1.5x \\
        MobileNetV3-Large-0.75 \pcite{howard2019searching} & 73.3\% & 4.0M & 1.5x \\
        FBNet-A \pcite{wu2019fbnet} & 73.0\% & 4.3M & 1.8x \\
        
        \midrule
        \textbf{\fcell{}-C} \; ($\lambda=3 \times 10^{-7}$)  & \textbf{77.1$\pm$0.03\%} & \textbf{4.4$\pm$0.04M} & 1.0x \\
        AtomNAS-B \pcite{mei2020atomnas} & 75.5\% & 4.4M & 1.0x \\
        FBNet-B \pcite{wu2019fbnet} & 74.1\% & 4.5M & 1.0x \\
        MnasNet-A2 \pcite{tan2019mnasnet} & 75.6\% & 4.8M & 1.1x \\
        One-Shot-Small \pcite{bender2018understanding} & 74.2\% & 5.1M & 1.2x \\
        MobileNetV2-1.3x \pcite{sandler2018mobilenetv2} & 74.4\% & 5.3M & 1.2x \\
        PC-DARTS \pcite{xu2020pc} & 75.8\% & 5.3M & 1.2x \\
        EfficientNet-B0 \pcite{tan2019efficientnet} & 76.3\% & 5.3M & 1.2x \\
        MobileNetV3-Large-1.0 \pcite{howard2019searching} & 75.2\% & 5.4M & 1.2x \\
        FBNet-C \pcite{wu2019fbnet} & 74.9\% & 5.5M & 1.3x \\
        

        \bottomrule[0.15em]
        \end{tabular}   
    }
    \caption{Extended version of Table 1: Comparison with modern mobile classification architectures and DNAS methods on ImageNet. FiGS produces the smallest and most accurate models in each category. Ratio-to-Ours indicates how much larger each network is compared to ours.}
    \label{imagenet-table}
\end{table*}



\end{document}